\documentclass[10pt,twocolumn,letterpaper]{article}
\usepackage{multirow}
\usepackage{subfig}
\usepackage{cvpr}
\usepackage{times}
\usepackage{epsfig}
\usepackage{graphicx}
\usepackage{amsmath}
\usepackage{array}
\usepackage{amssymb}

\usepackage[pagebackref=true,breaklinks=true,letterpaper=true,colorlinks,bookmarks=false]{hyperref}

\cvprfinalcopy 

\def\cvprPaperID{24} 

\ifcvprfinal\fi
\begin{document}
	
	\title{Facial Affect Estimation in the Wild Using Deep Residual and Convolutional Networks}
	
\author{Behzad Hasani and Mohammad H. Mahoor\\
	Department of Electrical and Computer Engineering	\\
	University of Denver, Denver, CO\\
	{\tt\small behzad.hasani@du.edu and mmahoor@du.edu}
	}
	
	\maketitle
	
	\begin{abstract}

		Automated affective computing in the wild is a challenging task in the field of computer vision. This paper presents three neural network-based methods proposed for the task of facial affect estimation submitted to the First Affect-in-the-Wild challenge. These methods are based on Inception-ResNet modules redesigned specifically for the task of facial affect estimation. These methods are: Shallow Inception-ResNet, Deep Inception-ResNet, and Inception-ResNet with LSTMs. These networks extract facial features in different scales and simultaneously estimate both the valence and arousal in each frame. Root Mean Square Error (RMSE) rates of 0.4 and 0.3 are achieved for the valence and arousal respectively with corresponding Concordance Correlation Coefficient (CCC) rates of 0.04 and 0.29 using Deep Inception-ResNet method.

	\end{abstract}
	
	\section{Introduction}
	
	Affect is a psychological term for describing the external exhibition of internal emotions and feelings. Affective
	computing attempts to develop systems that can interpret and estimate human affects
	through different channels (\eg visual, auditory, biological signals, etc.)~\cite{tao2005affective}. Facial  expressions  are  one  of  the  primary  non-verbal  communication methods for expressing emotions and intentions.
	
	There have been numerous studies for developing reliable automated Facial Expression Recognition (FER)
	systems in the past. However, current available systems are still far from desirable emotion perception
	capabilities required for developing robust and reliable Human Machine Interaction (HMI) systems. This is predominantly because of the fact that these HMI systems are needed to be in an uncontrolled environment (aka wild setting)
	where  there are significant  variations in the lighting, background, view, subjects' head pose, gender, and ethnicity~\cite{c53}. 
	
	Three models of Categorical, Dimensional, and FACS are proposed in the literature to quantify affective facial behaviors.
		In categorical model, emotion is chosen from a list of affective-related categories such as six basic emotions (anger, disgust, fear, happiness, sadness, and surprise) defined by Ekman \etal~\cite{c5}.
		In Dimensional model, a value is assigned to an emotion over a continuous emotional
		scale, such as ``valence" and ``arousal" defined in~\cite{russell1980circumplex}.
		In Facial Action Coding System (FACS) model, all possible
		facial component actions are described in terms of Action Units (AUs)~\cite{ekman1977facial}. FACS model only describes facial movements and
		does not interpret the affective state directly.
	
	The dimensional modeling of affect can distinguish between subtle differences in exhibiting of affect
	and encode small changes in the intensity of each emotion on a continuous scale, such as \textit{valence} and  \textit{arousal} where
	valence shows how positive or negative an emotion is, and arousal indicates how much an event is intriguing/agitating or calming/soothing~\cite{russell1980circumplex}. The First Affect-in-the-Wild challenge, focuses on estimation of valence and arousal (dimentional model of affect) in the wild.

	In this paper, we propose different methods submitted to the First Affect-in-the-Wild challenge for estimating dimensional model values of affect (valence and arousal) in the wild using deep convolutional networks and Long Short-Term Memory (LSTM) units. We also report the results of our methods on the Aff-Wild database provided in the First Affect-in-the-Wild challenge.

		The remainder of the paper is organized as follows: Section \ref{sec:2}  provides an overview of the related work in this field. Section \ref{sec:3} explains the methods submitted to the challenge. Experimental results and their analysis are presented in Section~\ref{sec:4} and finally the paper is concluded in Section \ref{sec:6}. 
	
	\section{Related work}\label{sec:2}
	
	  Conventional algorithms for affective computing from faces use different engineered features such as Local Binary Patterns (LBP)~\cite{c2}, Histogram of Oriented Gradients (HOG)~\cite{c14,7869885},  Histogram of Optical Flow (HOF)~\cite{c15}, and facial landmarks~\cite{c11,c12}. These features often lack required generalizability in cases where there is high variation in lighting, views, resolution, subjects' ethnicity, etc. Also, most of these works are applied on the categorical model of affect which can be considered an easier task than the dimensional model estimation task.
	  
	  Few number of studies have been conducted on the dimensional model of affect in the literature. Nicolaou \etal~\cite{nicolaou2011continuous}  trained bidirectional LSTM on multiple engineered features extracted from audio, facial geometry, and shoulders. They achieved Root Mean Square Error (RMSE) of 0.15 and Correlation Coefficient (CC) of 0.79 for valence as well as RMSE of 0.21 and CC of  0.64 for arousal.
	  
	  He \etal~\cite{he2015multimodal} won the AVEC 2015 challenge by training multiple stacks of bidirectional LSTMs (DBLSTM-RNN) on engineered features extracted from audio (LLDs features), video (LPQ-TOP features), 52 ECG features, and 22 EDA features. They achieved RMSE of 0.104 and CC of 0.616 for valence as well as RMSE of 0.121 and CC of 0.753  for  arousal.
	  
	  Koelstra \etal~\cite{koelstra2012deap} trained Gaussian naive Bayes classifiers  on EEG, physiological signals, and multimedia features by binary classification of low/high categories for arousal, valence, and liking on their proposed database DEAP. They achieved F1-score of 0.39, 0.37, and 0.40 on arousal, valence, and Liking categories respectively.

	In recent years, Convolutional Neural Networks (CNNs) have become the most popular  approach  among  researchers in the field of computer vision.
	Szegedy \emph{et al.} \cite{c18} introduced GoogLeNet which contains  multiple  ``Inception"  layers that apply several convolutions on the feature map in different scales. Several variations of Inception have been proposed~\cite{c20,c19}. Also, Inception layer is combined with residual unit introduced by He \emph{et al.}~\cite{c6} resulting considerable acceleration in the training of Inception networks~\cite{c23}.

	Recurrent Neural Networks (RNNs) can learn temporal dynamics by mapping input sequences to a sequence of hidden states~\cite{donahue2015long}. One of the problems of RNNs is that it is difficult for them to learn long-term sequences. This is mainly due to the vanishing or exploding gradients problem~\cite{C60}. LSTMs~\cite{C60} contain  a memory unit which solves this problem by  memorizing the context information for long periods of time. LSTM modules have three gates: 1) the input gate $(i)$ 2) the forget gate $(f)$ and 3) the output gate $(o)$ which overwrite, keep, or retrieve the memory cell $c$ respectively at the timestep $t$. Letting $ \sigma$ be the sigmoid function, $ \phi$ be the hyperbolic tangent function, and $\circ$ denoting Hadamard product, the LSTM updates for the timestep $t$ given inputs $x_t$, $h_{t-1}$, and $c_{t-1}$ are as follows:

	\begin{equation}
	\begin{split}
	&f_t = \sigma (W_f\cdot[h_{t-1},x_t] + b_f) \\
	&i_t = \sigma (W_i\cdot[h_{t-1},x_t] + b_i) \\
	&o_t = \sigma (W_o\cdot[h_{t-1},x_t] + b_o) \\
	&g_t =  \phi (W_C\cdot[h_{t-1},x_t] + b_C)\\
	&C_t = f_t \circ C_{t-1} + i_t \circ g_t \\
	&h_t = o_t \circ \phi ( C_t )
	\end{split}
	\label{eq:LSTM}
	\end{equation}
		
	Several works have used LSTMs and their extensions for different tasks. Fan \emph{et al.}~\cite{Fan:2016}  won the EmotiW 2016 challenge by cascading 2D-CNN with LSTMs and combining the resulting feature map with 3D-CNNs for facial expression recognition task. Donahue \etal~\cite{donahue2015long} proposed  Long-term  Recurrent  Convolutional  Network (LRCN) by combining CNNs and LSTMs. Byeon \emph{et al.}~\cite{byeon2015scene} proposed an LSTM-based network by applying 2D-LSTMs in four direction sliding windows. As mentioned earlier, Nicolaou \etal~\cite{nicolaou2011continuous} used bidirectional LSTMs and He \etal~\cite{he2015multimodal} used  multiple stacks of bidirectional LSTMs (DBLSTM-RNN) for the dimensional model of affect.
	
	In order to evaluate our methods, we calculate and report Root Mean Square Error (RMSE), Correlation Coefficient (CC), Concordance Correlation Coefficient (CCC), and Sign Agreement Metric (SAGR) metrics for our methods. In the following, we briefly review the definitions of these metrics.
	
	Root Mean Square Error (RMSE) is the most common evaluation metric in a continuous domain which is
	defined as:
	
	\begin{equation}
	RMSE = \sqrt{\frac{1}{n} \sum_{i=1}^{n}(\hat{\theta}_i-\theta_i)^2}
	\end{equation}
	where $\hat{\theta}_i$ and $\theta_i$ are the prediction and the ground-truth of $i^{\text{th}}$ sample, and $n$ is the number of samples. RMSE-based evaluation metrics can heavily weigh the outliers~\cite{bermejo2001oriented}, and they do not consider covariance of the data. 
	
	Pearson's Correlation Coefficient (CC) overcomes this problem~\cite{nicolaou2011continuous, schuller2011avec, schuller2012avec} and it is defined as: 
	
	\begin{equation}
	CC = \frac{COV\{\hat{\theta}, \theta\}}{\sigma_{\hat{\theta}}\sigma_{\theta}} = 
	\frac{E [(\hat{\theta}-\mu_{\hat{\theta}})(\theta-\mu_{\theta})]}{\sigma_{\hat{\theta}}\sigma_{\theta}}
	\end{equation}
	Concordance Correlation Coefficient (CCC) is another metric~\cite{ringeval2015avec, valstar2016avec} which combines CC with the square difference between the means of two compared time series:
	
	\begin{equation}
	\rho_c = \frac{2\rho \sigma_{\hat{\theta}} \sigma_{\theta}}{\sigma_{\hat{\theta}}^2 + \sigma_{\theta}^2 + (\mu_{\hat{\theta}} - \mu_\theta)^2}
	\end{equation}
	where $COV$ is covariance function, $\rho$ is the Pearson correlation coefficient (CC) between two time-series (e.g., prediction and ground-truth), $\sigma_{\hat{\theta}}^2$ and $\sigma_{\theta}^2$ are the variance of each time series, $\sigma_{\hat{\theta}}$ and $\sigma_{\theta}$ are the standard deviation of each, and $\mu_{\hat{\theta}}$ and $\mu_{\theta}$ are the mean value of each. Unlike CC, the predictions that are well correlated with the ground-truth but shifted in value are penalized in proportion to the deviation in CCC.

	The value of valence and arousal fall within the interval of [-1,+1] and correctly predicting their signs are essential in many emotion-prediction applications. Therefore, we use Sign Agreement Metric (SAGR) which is proposed in~\cite{nicolaou2011continuous} to evaluate the performance of a valence and arousal prediction system with respect to the sign agreement. SAGR is defined as:
	\begin{equation}
	SAGR = \frac{1}{n}\sum_{i=1}^{n}\delta{(sign(\hat{\theta}_i),sign(\theta_i))}
	\end{equation}
	where $\delta$ is the Kronecker delta function, defined as:
	\begin{equation}
	\delta{(a,b)} = 
	\begin{cases}
	1,				& a = b\\
	0,              & a \neq b
	\end{cases}
	\end{equation}   
	

	\section{Proposed methods}\label{sec:3}
	
	Inception and ResNet have shown remarkable results in various tasks~\cite{behzad2017,c4,c18,Yao:2016:HTR:2993148.2997639}. For the Affect-in-the-Wild challenge, we proposed Inception-ResNet based architectures followed by LSTM units (submission~3) for the task of affect estimation.  Our proposed methods extract contextual information of the frames in an end-to-end deep neural network. In the following, we explain each of the methods presented in our submissions.   
		\begin{figure}[!tbp]
			\begin{center}
					\includegraphics[width=\linewidth]{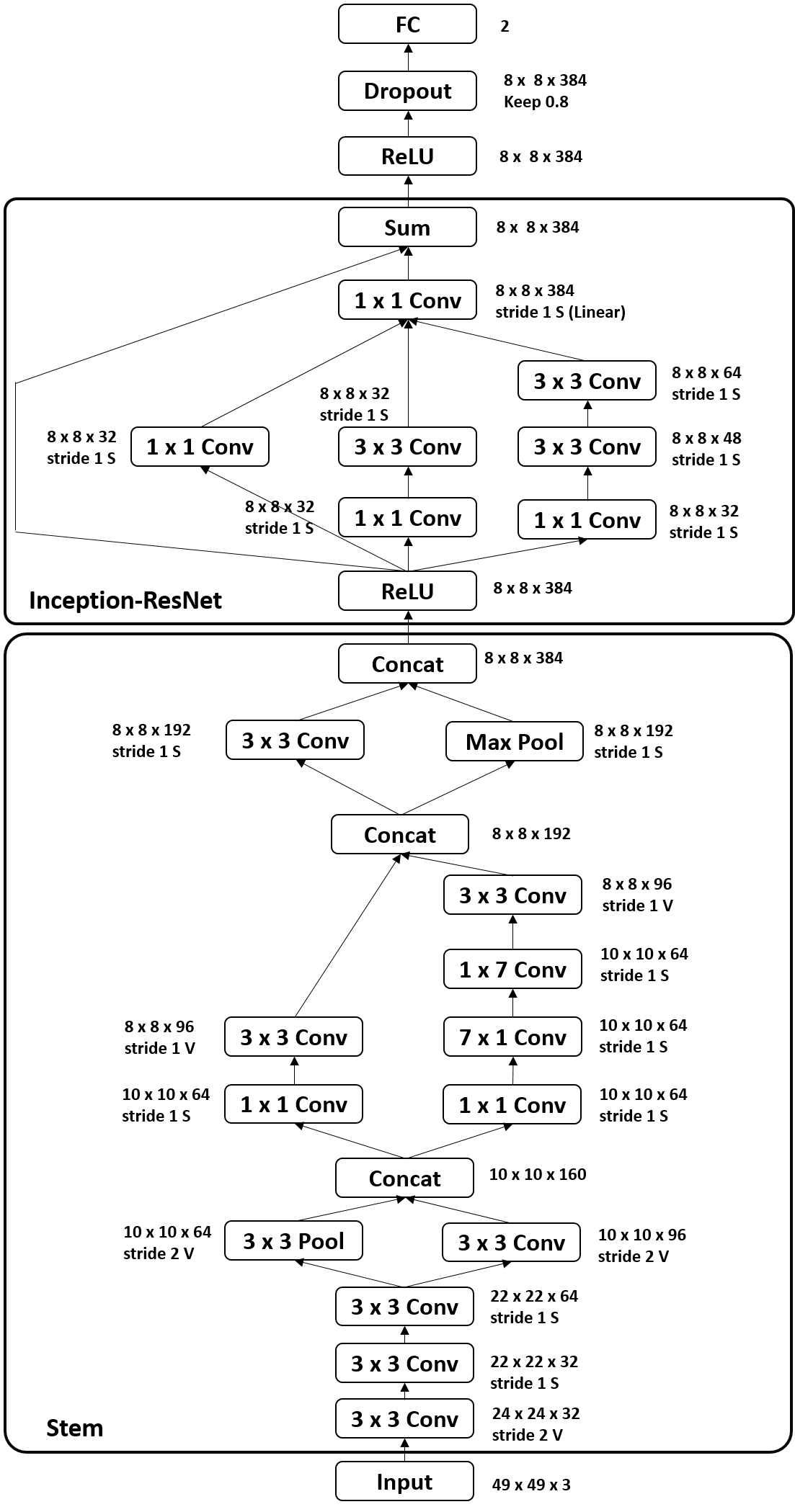}
			\end{center}
			\caption{Network architecture for submission 1. The ``V" and ``S" marked layers represent ``Valid" and ``Same" paddings respectively. The size of the output tensor is provided next to each layer.}
			\label{fig:sub1}
		\end{figure}

	\subsection{Shallow Inception-ResNet (submission~1)}
	
	For the first submission, we propose modified version of Inception-ResNet which originally presented in~\cite{c23}. Our first module  is  shallower  than the original Inception-ResNet containing only ``stem'' and single ``Inception-ResNet" module. Most of the settings are the same as the ones presented in~\cite{c23} while the input size of the network is changed from $299\times299$ to $49\times49$. Because of this reduction in the size of the input, we are not able to have a very deep network. Therefore, only one Inception-ResNet module is used in  this method.
	
	Figure \ref{fig:sub1} shows the structure of our shallow Inception-ResNet method. The input images with the size $49\times49\times3$ are followed by the ``stem" layer. Afterwards, the stem is followed by Inception-ResNet-A, dropout, and a fully-connected layer respectively. In Figure \ref{fig:sub1}, detailed specification of each layer is provided. All convolution layers are followed by a batch normalization layer and all batch normalization layers (except the ones that are indicated as ``Linear" in Figure \ref{fig:sub1}) are followed by a ReLU \cite{c21} activation function to avoid the vanishing gradient problem.

	\begin{figure}[!tbp]
		\begin{center}
			\includegraphics[width=0.95\linewidth]{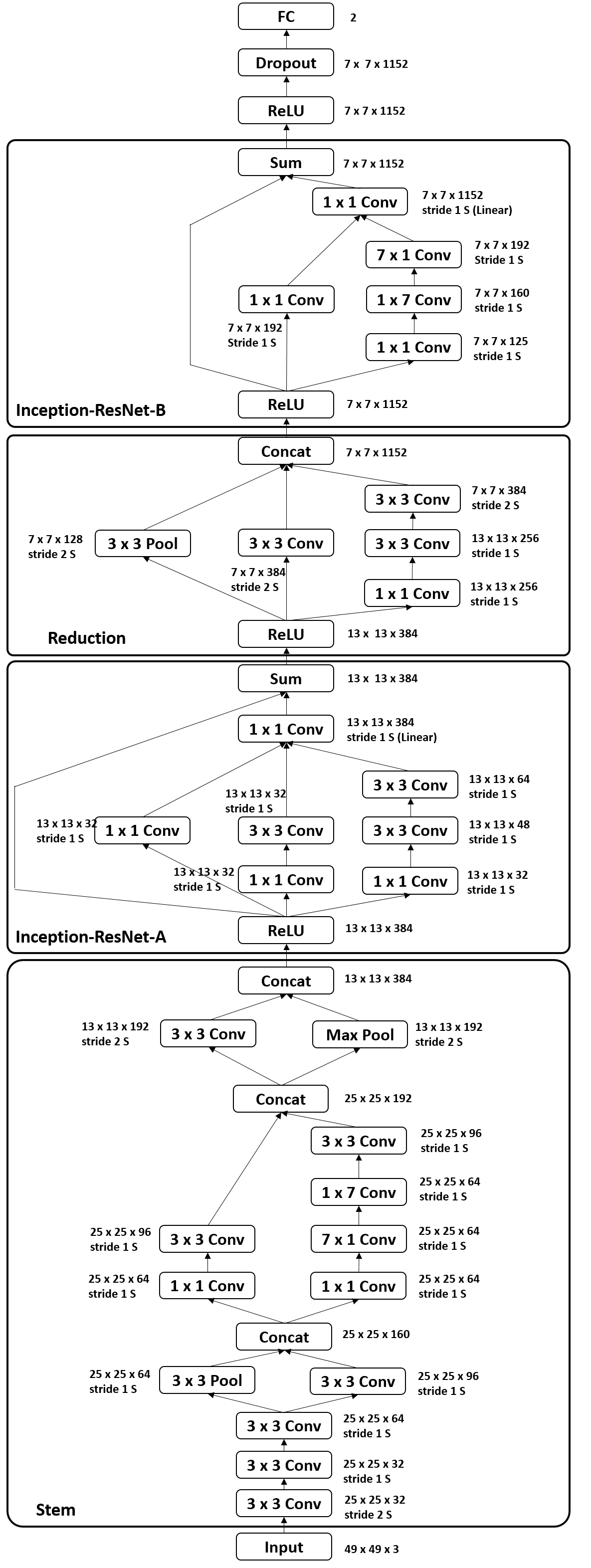}
		\end{center}
		\caption{Network architecture for submission 2. The ``S" marked layers represent ``Same" padding. The size of the output tensor is provided next to each layer.}
		\label{fig:sub2}
	\end{figure}

    \subsection{Deep Inception-ResNet (submission~2)}
	As mentioned earlier, in order to have a deeper network to extract more abstract features and having more number of parameters to learn, we changed the properties of the previously mentioned network.  
	
	Figure \ref{fig:sub2} shows the structure of our Deep Inception-ResNet method. Comparing tho the previous method, we change all of the ``valid" paddings to ``same" paddings to save the feature map size. Also, strides are changed in this method. Same as before, the input images with the size $49\times49\times3$ are followed by the ``stem" and ``Inception-ResNet-A" layers. Afterwards, to deepen the network, we include ``Reduction" (which reduces the grid size from $13\times13$ to $7\times7$) followed by ``Inception-ResNet-B", dropout, and fully-connected layers.
	
	Same as before, all convolution layers are followed by a batch normalization layer and all batch normalization layers (except the ones that are indicated as ``Linear" in the Figure) are followed by a ReLU activation function to avoid the vanishing gradient problem. 
	
	\subsection{Inception-ResNet \& LSTM (submission~3)}
	As explained earlier, LSTMs have shown remarkable results in different emotion recognition/estimation tasks~\cite{donahue2015long,Fan:2016,he2015multimodal,nicolaou2011continuous}. Therefore, we incorporate LSTMs in our next method in two directions to estimate the valence and arousal intensity in the challenge.
	
\begin{figure}[!tbp]
		\begin{center}
			\includegraphics[width=0.89\linewidth]{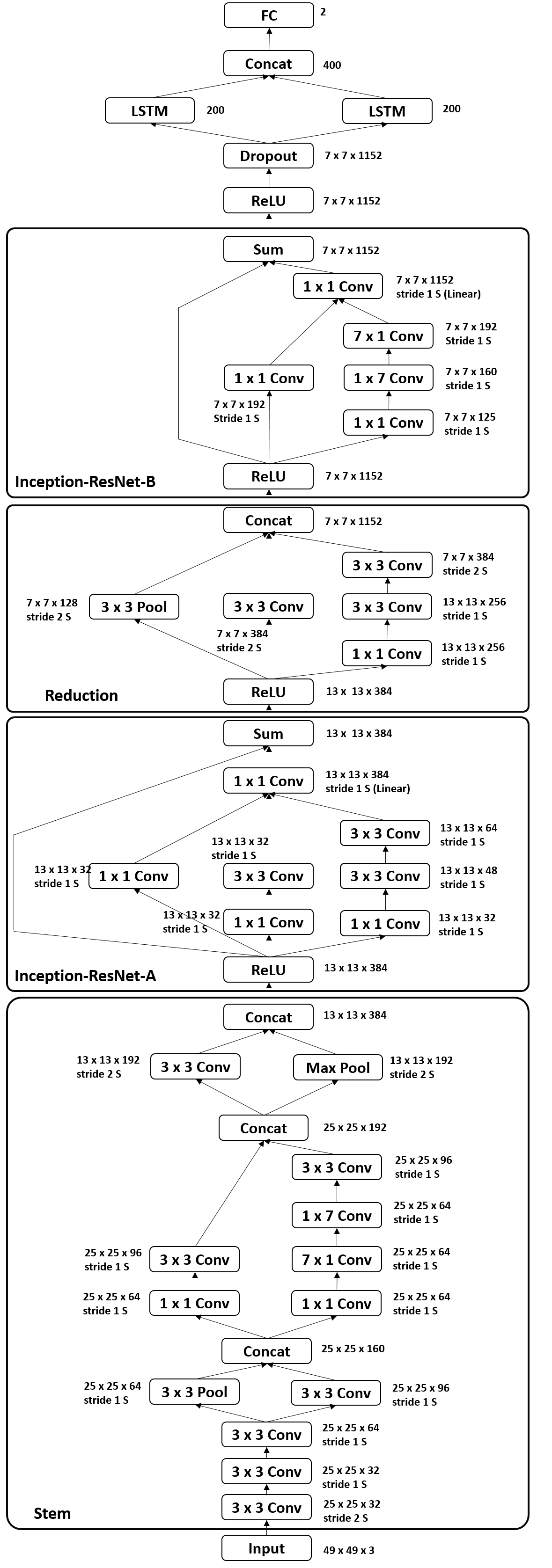}
		\end{center}
		\caption{Network architecture for submission 3. The ``S" marked layers represent ``Same" padding. The size of the output tensor is provided next to each layer.}
		\label{fig:sub3}
	\end{figure}

	 Figure \ref{fig:sub3} shows the network used for the third submission. The network has the same settings as the second submission. The only difference here is that after the dropout layer, we vectorize the feature map on two dimensions (one on the width of the feature map and the other one on its height). This is inspired by the work in~\cite{byeon2015scene} where LSTMs are used in four different directions. Adding LSTMs will take the complex spatial dependencies of adjacent pixels into account~\cite{byeon2015scene}.  We investigated that 200 hidden units for each LSTM unit is a reasonable amount for this task. At the end, the resulting feature vectors of these two LSTMs are concatenated together and are followed by a fully-connected layer (Figure~\ref{fig:sub3}).
	 
	All of the  proposed  methods  are  implemented  using a combination of TensorFlow~\cite{abadi2016tensorflow} and TFlearn~\cite{tflearn2016} toolboxes on NVIDIA Tesla K40 GPUs. We used asynchronous stochastic gradient descent with weight decay of 0.0001, and learning rate of 0.01. Mean square error used for loss function.
	
	\section{Database \& results}\label{sec:4}
	
	In this section, we briefly review Aff-Wild database provided  in the First Affect-in-the-Wild challenge. We then report the results of our experiments on both validation and test sets using the metrics provided in section~\ref{sec:2}.
		\begin{figure*}[!tbp]
			\centering
			\subfloat[Annotated values (ground-truth)]{\includegraphics[width=0.35\textwidth]{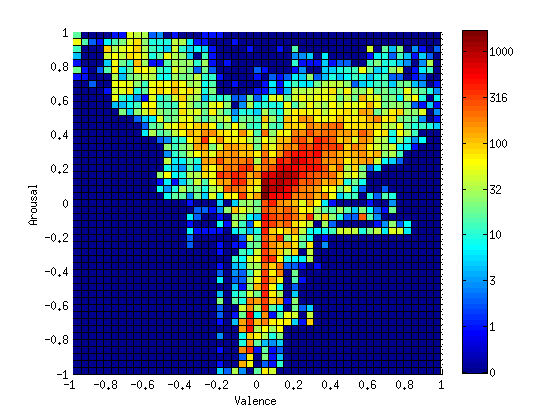}\label{fig:GT}}
			\hspace{1em}
			\subfloat[Shallow Inception-ResNet (submission 1)]{\includegraphics[width=0.35\textwidth]{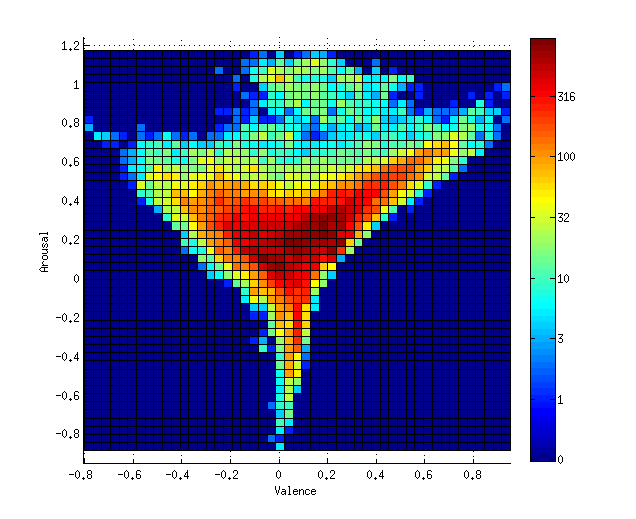}\label{fig:sub1_result}}
			\hspace{1em}\\
			\subfloat[Deep Inception-ResNet (submission 2)]{\includegraphics[width=0.35\textwidth]{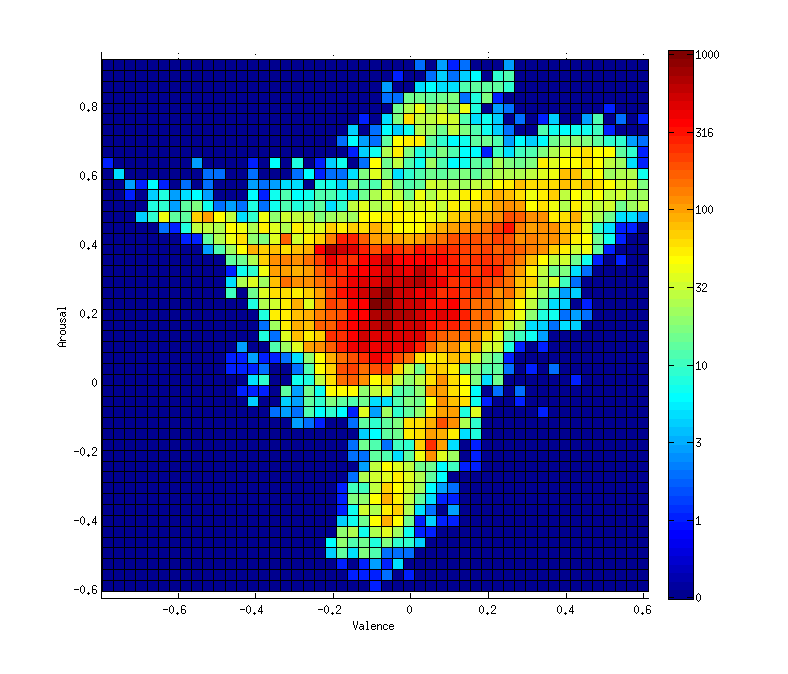}\label{fig:sub2_result}}
			\hspace{1em}
			\subfloat[Inception-ResNet with LSTMs (submission 3)]{\includegraphics[width=0.35\textwidth]{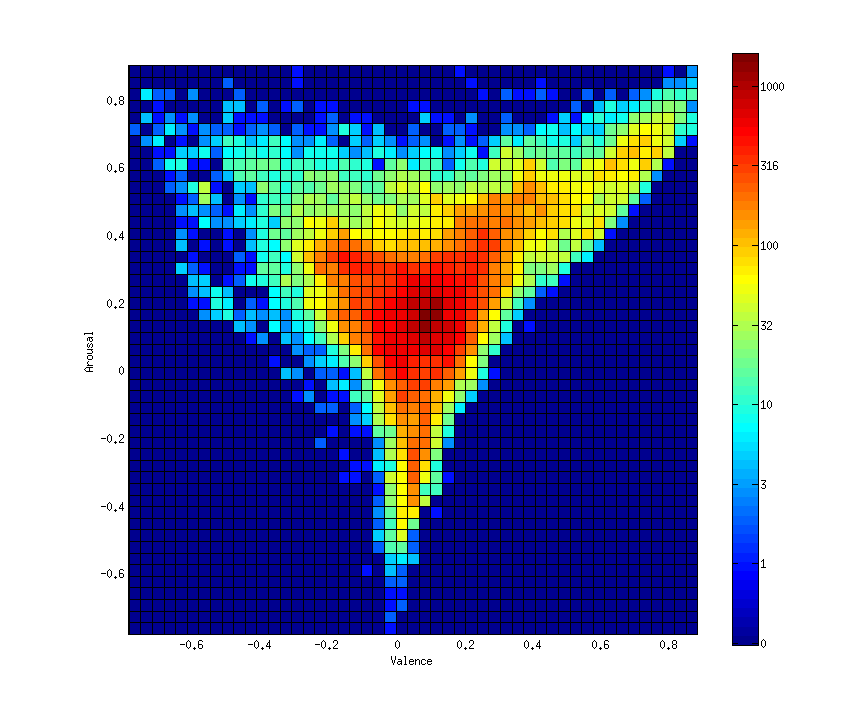}\label{fig:sub3_result}}
			\caption{Histograms of valence and arousal values in the validation set for annotated values (a), submission 1 (b), submission 2 (c), and submission 3 (d). Best viewed in color. }
			\label{fig:histograms}
		\end{figure*}
	\subsection{Aff-Wild database}
	Aff-Wild database contains 300 videos  of different subjects watching videos of various TV shows and movies. The videos contain subjects from different genders and ethnicities with high variations in head pose and lightning. Provided videos in this database are  annotated with  valence and arousal values for each frame. 254 videos of this database are selected for training and the rest 46 videos are used for evaluating the participants in the challenge.

	\subsection{Results}
	
    We first extract the faces from the frames using the bounding boxes that are released alongside the database. Afterwards, we resize the faces to $49\times49$ pixels and divide the training data into training and validation sets by assigning 10 percent of the subjects to the validation set and the rest of them to the training set.

	Figure~\ref{fig:histograms} shows the histograms of annotated values and predicted values on the validation set for all submissions. It can be seen that the database is heavily imbalanced. The number of annotated frames with values close to zero for valence and arousal (center of the circumplex) is considerably higher than other regions. Therefore, our methods are also biased toward this region. Figures~\ref{fig:sub1_result} and~\ref{fig:sub2_result} show that submissions 1 and 2 were not able to correctly estimate instances with high arousal and low valence values. Figure~\ref{fig:sub3_result} shows that submission 3 performs better in this region but as table~\ref{Results} indicates, this submission generally does not perform well on estimating arousal values comparing to others. However, all of the methods show mostly similar patterns on the validation set and no unusual predicted values can be seen throughout any of the methods.


\newcolumntype{?}{!{\vrule width 1.5pt}}

\begin{table*}[!]
	\begin{center}
		\begin{tabular}{|c?c|c|c|c|c|c|c|c?c|c|c|c|}
		\hline
		\multirow{3}{*}{submission} & \multicolumn{8}{c?}{validation set}                                                                        & \multicolumn{4}{c|}{test set}                        \\ \cline{2-13} 
		                            & \multicolumn{2}{c|}{RMSE} & \multicolumn{2}{c|}{CC} & \multicolumn{2}{c|}{CCC} & \multicolumn{2}{c?}{SAGR} & \multicolumn{2}{c|}{RMSE} & \multicolumn{2}{c|}{CCC} \\ \cline{2-13} 
		                            & valence     & arousal     & valence    & arousal    & valence     & arousal    & valence     & arousal     & valence     & arousal     & valence     & arousal    \\ \hline
		\#1                         & 0.29        & 0.37        & 0.29       & 0.22       & 0.28        & 0.19       & 0.53        & 0.72        & 0.41        & 0.33        & 0.03        & 0.19       \\ \hline
		\#2                         & 0.27        & 0.36        & 0.44       & 0.26       & 0.36        & 0.19       & 0.57        & 0.74        & 0.40        & 0.30        & 0.04        & 0.29       \\ \hline
		\#3                         & 0.28        & 0.36        & 0.33       & 0.17       & 0.31        & 0.14       & 0.55        & 0.70        & 0.40        & 0.33        & 0.04        & 0.17       \\ \hline
		\end{tabular}
	\end{center}
	\caption{Results of submissions on validation and test sets }
	\label{Results}
\end{table*}
	We evaluate our proposed methods with RMSE, CC, CCC, and SAGR metrics defined in the section~\ref{sec:2}. The reported results are the calculated values on the validation set.
	Table~\ref{Results} shows different metrics calculated for the validation and test sets of the provided database. The CC and SAGR metric are not reported to the authors for the test set, therefore these metrics are not reported in Table~\ref{Results}. 
	
	Results on validation set show more accurate estimation for arousal in all submissions. This can be in part due to the less dynamic range of values for arousal in the training data.  By looking at the results on our validation set, it can be seen that almost all of the metrics show the superiority of  submission~2 comparing to other submissions. CC and CCC metrics show that there is more correlation between the results of submission~2 and the ground-truth comparing to other methods.
	
	The test set results in Table~\ref{Results} also show the superiority of submission~2. In all three methods, the estimation for valence is considerably less accurate comparing to the estimated values for arousal while the results in the validation set do not show such drastic difference. Nevertheless, submissions~1~and~3 show almost the same results on the test set while submission~2 shows less error in terms of RMSE and also shows more correlation with ground-truth in terms of CCC metric. The reduction of correlation in submission 3 can be in part due to the fact that the input feature map of LSTM units does not contain the notion of time which shapes an unfitting input for the LSTMs. Using 3D convolutional neural networks would provide such temporal information within the feature map but this temporal processing of input sequences is not experimented in this work.

	
	\section{Conclusion}\label{sec:6}
	In this paper, we presented three methods submitted to the First Affect-in-the-Wild Challenge: Shallow Inception-ResNet, Deep Inception-ResNet, and Inception-ResNet with LSTMs. These Inception-ResNet based methods are engineered specifically for the task of facial affect estimation by extracting facial features in different scales and they estimate both valence and arousal values for each frame simultaneously. We used four metrics to evaluate our methods on our validation set: RMSE, CC, CCC, and SAGR. On the test set, Inception-ResNet with LSTMs network achieved the best performance with noticeably good estimation in terms of RMSE and CCC  rates especially on arousal values.

	\section{Acknowledgement}
	
	This work is partially supported by the NSF grants IIS-1111568 and CNS-1427872. 
	We gratefully acknowledge the support from NVIDIA Corporation with the donation of the Tesla K40 GPUs used for this research.

	
	{\small
		\bibliographystyle{ieee}
		\bibliography{egbib}
	}
	
\end{document}